\documentclass[letterpaper]{article} 
\usepackage{aaai2026}  
\usepackage{times}  
\usepackage{helvet}  
\usepackage{courier}  
\usepackage[hyphens]{url}  
\usepackage{graphicx} 
\usepackage{natbib}  
\usepackage{caption} 
\usepackage{algorithm}
\usepackage{algorithmic}

\usepackage{newfloat}
\usepackage{listings}
\DeclareCaptionStyle{ruled}{labelfont=normalfont,labelsep=colon,strut=off} 
\floatstyle{ruled}
\newfloat{listing}{tb}{lst}{}
\floatname{listing}{Listing}
\usepackage[utf8]{inputenc} 
\usepackage{url}            
\usepackage{booktabs}       
\usepackage{amsfonts}       
\usepackage{amsmath}
\usepackage{nicefrac}       
\usepackage{microtype}      

\usepackage{cleveref}
\usepackage{graphicx}
\usepackage{multirow}
\usepackage{makecell}
\usepackage{pifont}
\usepackage{xspace}

\newcommand{\modelName}{DriveSuprim\xspace}
\newcommand{\CM}{\ding{51}}
\newcommand{\CrossM}{\ding{55}}

\title{DriveSuprim: Towards Precise Trajectory Selection for End-to-End Planning}
\author{
    Wenhao Yao\textsuperscript{\rm 1, \rm 2}, Zhenxin Li\textsuperscript{\rm 1, \rm 2}, Shiyi Lan\textsuperscript{\rm 3}, Zi Wang\textsuperscript{\rm 3},\\ Xinglong Sun\textsuperscript{\rm 3},
    Jose M. Alvarez\textsuperscript{\rm 3}, Zuxuan Wu\textsuperscript{\rm 1, \rm 2}\thanks{Corresponding author.}
}
\affiliations{
    \textsuperscript{\rm 1}Shanghai Key Lab of Intell. Info. Processing, School of CS, Fudan University\\
    \textsuperscript{\rm 2}Shanghai Collaborative Innovation Center of Intelligent Visual Computing\\
    \textsuperscript{\rm 3}NVIDIA

    \{whyao23, lizx23\}@m.fudan.edu.cn, zxwu@fudan.edu.cn
}

\usepackage{bibentry}

\begin{document}

\maketitle

\begin{abstract}
Autonomous vehicles must navigate safely in complex driving environments. Imitating a single expert trajectory, as in regression-based approaches, usually does not explicitly assess the safety of the predicted trajectory. Selection-based methods address this by generating and scoring multiple trajectory candidates and predicting the safety score for each. However, they face optimization challenges in precisely selecting the best option from thousands of candidates and distinguishing subtle but safety-critical differences, especially in rare and challenging scenarios. We propose \textbf{\modelName} to overcome these challenges and advance the selection-based paradigm through a coarse-to-fine paradigm for progressive candidate filtering, a rotation-based augmentation method to improve robustness in out-of-distribution scenarios, and a self-distillation framework to stabilize training. \textbf{\modelName} achieves state-of-the-art performance, reaching 93.5\% PDMS in NAVSIM v1 and 87.1\% EPDMS in NAVSIM v2 without extra data, with 83.02 Driving Score and 60.00 Success Rate on the Bench2Drive benchmark, demonstrating superior planning capabilities in various driving scenarios.
\end{abstract}

\begin{links}
    \link{Code}{https://github.com/William-Yao-2000/DriveSuprim}
\end{links}

\section{Introduction}

End-to-end autonomous driving has traditionally relied on regression-based approaches that predict a single trajectory to mimic expert behavior
~\cite{jiang2023vad, hu2023planning, wang2023drivedreamer}
. While regression is a common approach, it fundamentally lacks the ability to evaluate multiple alternatives in safety-critical scenarios where subtle trajectory differences can significantly impact outcomes.

In recent years, selection-based methods~\cite{chen2024vadv2, li2024hydra, li2024hydramdp_pp, wang2025enhancing} have clearly outperformed regression approaches. The key is their capability to generate and evaluate diverse trajectory candidates using comprehensive safety metrics such as collision risk and driving rule compliance~\cite{dauner2024navsim}. This explicit comparison enables the system to select the safest and most appropriate trajectory from multiple alternatives, addressing safety-critical issues that regression-based methods cannot address. Our oracle study in Tab~\ref{tab:oracle} demonstrates the substantial potential of selection-based methods: when making ideal optimal selection, these approaches can even surpass human demonstrations in NAVSIM~\cite{dauner2024navsim} safety-critical metrics. This performance ceiling highlights why selection-based planning has become the preferred paradigm for autonomous driving systems requiring robust safety guarantees.

\begin{table}[!t]
    \centering
    \small
    \begin{tabular}{c|cccccc}
        \toprule
        Top-K & NC$\uparrow$ & DAC$\uparrow$ & EP$\uparrow$ & TTC$\uparrow$ & C$\uparrow$ & PDMS$\uparrow$ \\
        \midrule
        1 & 99.0 & 98.7 & 86.5 & 96.2 & 100 & 91.9 \\
        4 & 99.4 & 99.6 & 89.6 & 98.0 & 100 & 94.5 \\
        16 & 99.7 & 99.8 & 92.0 & 99.1 & 100 & 96.1 \\
        256 & 100 & 100 & 97.1 & 99.9 & 100 & 98.7 \\
        \midrule
        Human & 100 & 100 & 87.5 & 100 & 99.9 & 94.8 \\
        \bottomrule
    \end{tabular}
    \caption{PDM score of the best trajectory in the top-K candidates on ranked predicted scores.}
    \label{tab:oracle}
\end{table}

\begin{figure*}[t] \centering
    \includegraphics[width=0.95\textwidth]{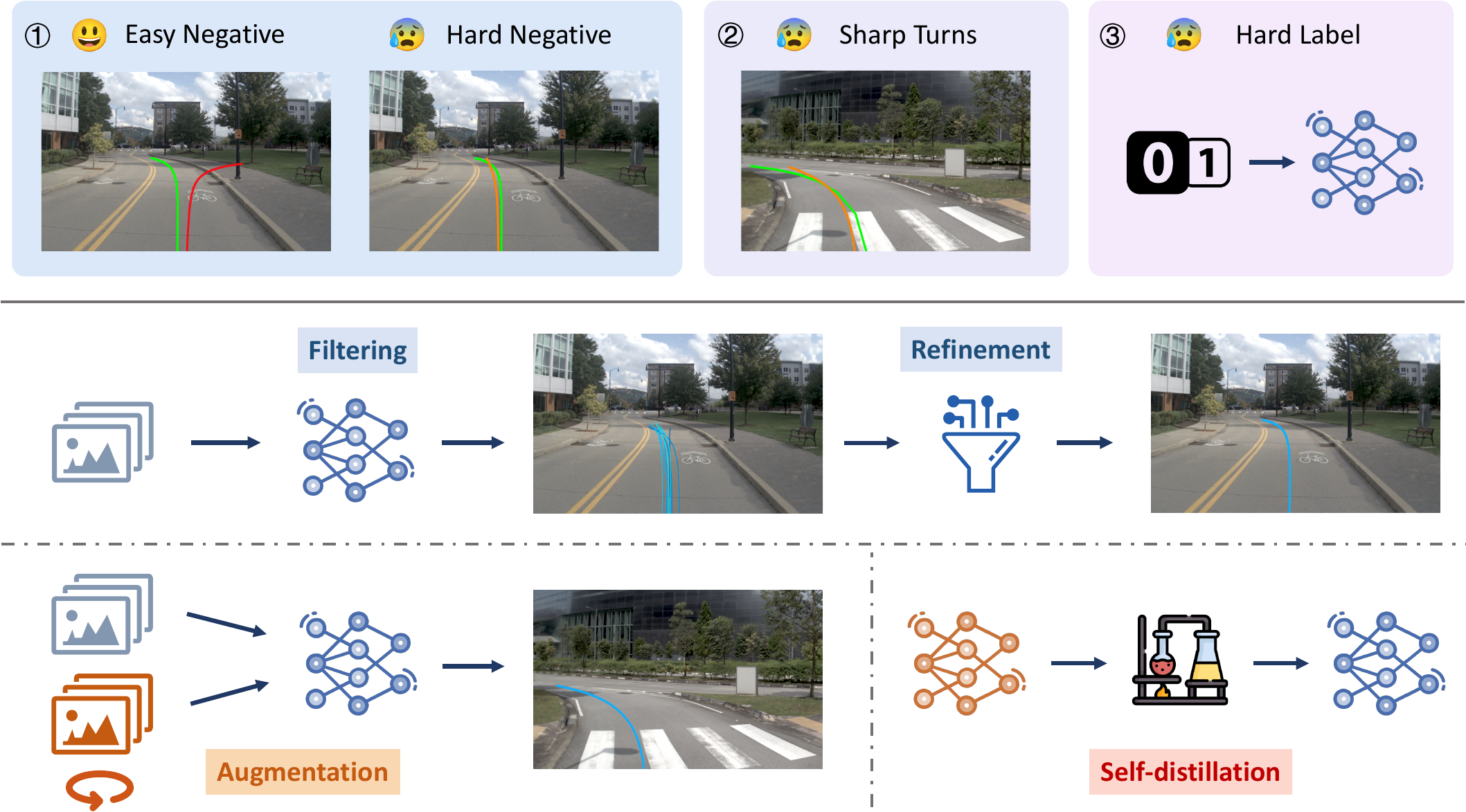}
    \caption{Overall pipeline of our method. 
    Selection-based methods struggle to distinguish suboptimal trajectories, perform poorly in turning, and utilize hard binary labels in training. 
    \modelName introduces coarse-to-fine refinement and a rotation-based data augmentation method with self-distillation to address these weaknesses. 
    The green trajectory is the ground-truth trajectory, and the red and orange trajectories are obviously unsafe and seemingly correct trajectory candidates in the trajectory vocabulary.
    }
    \label{fig:teaser}
\end{figure*}

However, selection-based methods still face three critical limitations, which make them difficult to reach the ideal perfect trajectory selection displayed in Tab~\ref{tab:oracle} and limit the model performance.
First, selection-based methods struggle to distinguish the optimal trajectory from similar but suboptimal alternatives. 
During training, the model encounters thousands of trajectory candidates where the vast majority are unsafe or impractical (``easy negatives'', the red trajectory in Fig~\ref{fig:teaser}). These easy-to-reject options dominate the training process and gradient, causing the model to focus primarily on avoiding obviously incorrect choices. In contrast, the model receives insufficient supervision to select the most suitable trajectory from reasonable-looking trajectories with subtle but important differences (``hard negatives'', the orange trajectory in Fig~\ref{fig:teaser}).
The overwhelming number of obvious negative examples hinders the model's ability to develop fine-grained discrimination capability, which is crucial for selecting optimal trajectories when presented with multiple plausible options, to avoid route deviation or collision.

Second, selection-based methods suffer from directional bias in trajectory distribution. This bias manifests as an imbalance in training data that, while reflecting real-world driving patterns where straight driving predominates, leads to models that perform relatively poorly in turning scenarios. Training with such imbalanced data naturally results in models that excel at straight-line driving but struggle with turns and complex maneuvers.
Even advanced autonomous driving datasets like NAVSIM exhibit this limitation.
We find that only 18\% of ground-truth trajectories in NAVSIM involve turns exceeding 30 degrees. While this distribution may reflect typical driving patterns, it creates a significant challenge for learning models, which require sufficient examples of all maneuver types to develop robust capabilities. This directional bias significantly impairs the model's ability to select the most precise large-angle turning trajectories, particularly in navigation-critical scenarios where turns are essential.

Third, selection-based methods typically rely on binary classification for safety-related decisions, labeling trajectories as ``safe'' or ``unsafe'' based on specific thresholds for collision risk or rule compliance. 
This binary approach creates hard decision boundaries where trajectories just above or below a safety threshold could be treated entirely differently. Slight variations in score could suddenly thoroughly flip a trajectory label from being selected to being rejected.
As a result, models risk becoming overly sensitive to minor changes in trajectory features, which causes inconsistent behavior.

We present \textbf{\modelName}, a novel method that tackles these three critical challenges in selection-based trajectory prediction and makes more nuanced and precise trajectory selection. 
Our contributions include:

\begin{itemize}

    \item We propose a coarse-to-fine refinement method that addresses the challenge of distinguishing between similar trajectories. Our method filters promising candidates and applies fine-grained scoring to the most challenging options, significantly improving discrimination between similar but subtly different trajectories.

    \item We propose an integrated training pipeline combining rotation-based data augmentation with self-distillation to address directional bias and hard decision boundaries. Our approach synthesizes challenging turning scenarios and leverages teacher-generated soft pseudo-labels, effectively balancing trajectory distributions and realizing better optimization for training a more robust model.

    \item \modelName achieves state-of-the-art performance on the NAVSIM and Bench2Drive benchmark, demonstrating the effectiveness of our model in handling challenging driving scenarios. \modelName significantly outperforms the previous methods by 3.6\% and 1.5\% on NAVSIMv1 and NAVSIMv2 without introducing other training data.
    On the Bench2Drive benchmark, our method achieves 83.02 and 60.00 on driving score and success rate.
\end{itemize}

\section{Related Works}
\subsection{End-to-end planning}
Autonomous driving has traditionally relied on modular pipelines that separate perception from planning. However, UniAD~\cite{hu2023planning} highlights several limitations of this approach, including information loss and error propagation. To address these challenges, end-to-end driving methods
~\cite{chen2020learning, chen2021learning, chitta2022transfuser, hu2023planning, jiang2023vad, li2024ego, li2024hydra, wang2025enhancing} 
unify the perception-to-planning pipeline within a single optimizable network. They process raw sensor inputs and directly output driving trajectories. While some methods
~\cite{chen2021learning, zhang2021end} 
use reinforcement learning (RL) to learn through interaction with simulated environments, the majority adopt imitation learning (IL), training from expert demonstrations without interaction. Most IL-based approaches
~\cite{chitta2022transfuser, hu2023planning, jiang2023vad, li2025finetuning, liao2024diffusiondrive} 
generate a single trajectory using regression or diffusion-based methods to mimic expert behavior.

More recently, selection-based methods~\cite{chen2024vadv2, li2024hydra, li2024hydramdp_pp, wang2025enhancing} have emerged. These models evaluate a diverse set of candidate trajectories by scoring them on safety-focused metrics (e.g., PDM scores~\cite{dauner2024navsim}). A prominent example is Hydra-MDP~\cite{li2024hydra}, which employs multiple rule-based teachers and distills them into the planner to create diverse trajectory candidates tailored to different evaluation metrics.

Our proposed model also falls under the selection-based paradigm. However, unlike prior methods that perform a single-shot selection from a fixed candidate set—probably leading to suboptimal decisions—we introduce a coarse-to-fine selection and refinement strategy. This approach significantly improves selection precision by progressively narrowing down the trajectory set to the most optimal candidates. 

\subsection{Iterative \& Multi-stage Refinement}
Iterative refinement has been widely adopted to improve results in optical flow
~\cite{ilg2017flownet,hui2018liteflownet,teed2020raft,xu2022gmflow}
and motion estimation
~\cite{sun2024refining,zheng2023pointodyssey}.
A common strategy in these works is to iteratively propagate features or trajectory estimates through a shared module to progressively refine the predictions. Inspired by this, iterative refinement is also applied in object detection~\cite{zhu2020deformable,cai2018cascade} to improve performance. For instance, in Deformable DETR, each decoder layer refines bounding boxes based on predictions from the previous layer.

We similarly adopt a multi-stage refinement strategy to improve trajectory selection accuracy. However, rather than repeatedly updating fixed-dimensional features, we implement selection from a fixed vocabulary of candidate trajectories. After selection, the search space is progressively narrowed to a more precise subset, improving the final prediction quality.

\subsection{Augmentation for Enhanced Robustness}
Robustness has long been a critical focus in computer vision research
~\cite{ganin2016domain, croce2020robustbench, rebuffi2021data, sun2022benchmarking}. 
Early studies demonstrated that image models are highly sensitive to minor domain shifts~\cite{azulay2019deep} and adversarial perturbations~\cite{szegedy2013intriguing, goodfellow2014explaining}.
For example, MNIST-C~\cite{mu2019mnist} introduces 15 distinct corruption types to benchmark model performance against diverse failure modes. Motivated by these insights, several methods
~\cite{rusak2020simple, mintun2021interaction, kar20223d} 
utilize corruption-based augmentations, such as adding
Gaussian and speckle noise, to enhance robustness. Inspired by these methods, our study explores the use of similar corruption-based augmentation techniques specifically for end-to-end driving models. We introduce targeted perturbations tailored to autonomous driving scenarios, addressing critical domain shifts—particularly the overrepresentation of straightforward driving trajectories—which pose challenges for scenarios involving complex maneuvers such as turns.

\begin{figure*}[!t]
    \centering
    \includegraphics[width=0.98\textwidth]{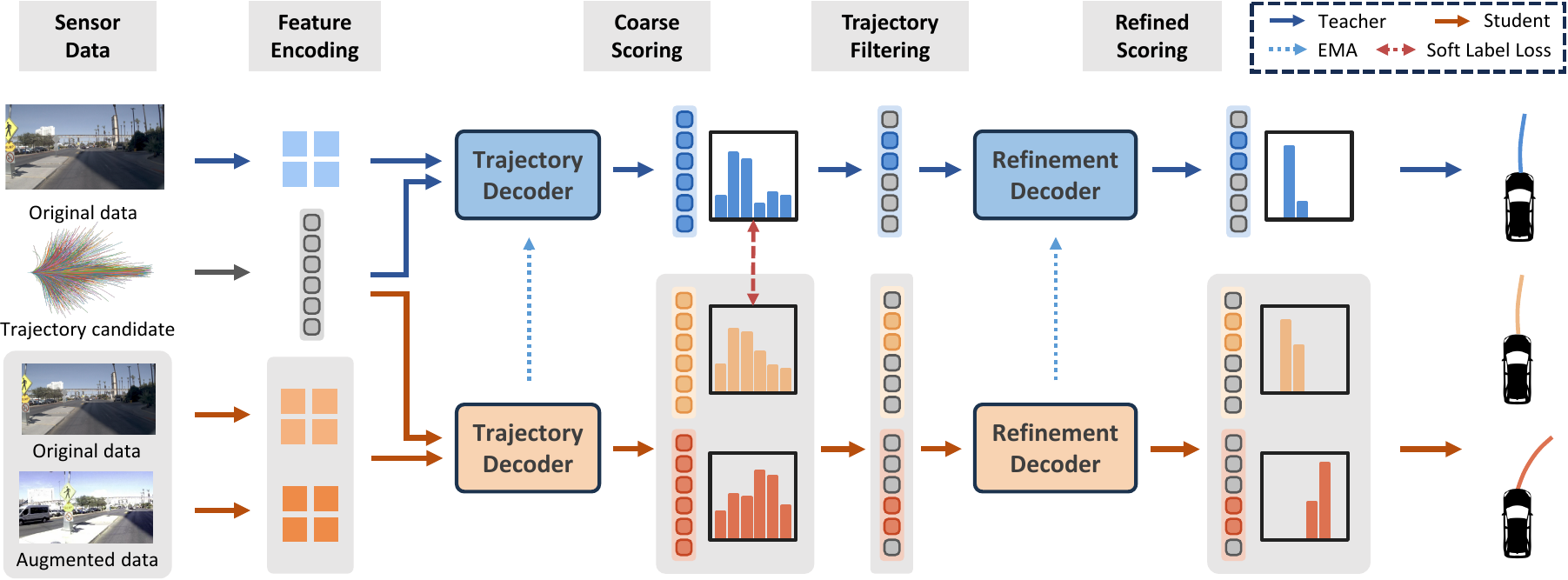}
    \caption{Model architecture. \modelName adopts a coarse-to-fine paradigm to better distinguish hard negatives. According to the scoring distribution, the Trajectory Decoder filters potential candidates, and the Refinement Decoder further outputs fine-grained trajectory scores. The model introduces rotation-based augmented data to ease the directional bias and applies a self-distillation framework for stable training. The teacher outputs serve as soft labels for auxiliary supervision for the student.
    }
    \label{fig:arch}
\end{figure*}

\section{Methods}

We introduce our method in this section. Firstly, we introduce preliminaries about the end-to-end planning and the selection-based method. Next, we introduce our proposed coarse-to-fine selection paradigm, rotation-based data augmentation method and self-distillation framework.

\subsection{Preliminaries}

\paragraph{End-to-End planning \& Selection-based planning}
In autonomous driving, the end-to-end planning requires the planning system to output a future trajectory $T$ based on input sensor data, like RGB image or Lidar point cloud:
\begin{equation}
    T = \mathrm{Planner}\left(Img, Lidar \right),
\end{equation}
where the trajectory $T$ can be represented as a sequence of vehicle locations $(u_1, u_2, ..., u_l)$ or a sequence of controller actions $(a_1, a_2, ..., a_l)$, and $l$ denotes the sequence length.

Among the end-to-end planning methods, the selection-based paradigm predefines a trajectory vocabulary $\{\tau_i\}_{i=1}^N$ covering $N$ planning trajectories. Given a specific driving scenario, the quality of each trajectory is evaluated by several metrics, like the $l_2$ distance to the human teacher trajectory, or metrics considering driving safety and traffic rule adherence. The model learns a scorer that generates trajectory scores $\{s_i\}_{i=1}^N$ revealing trajectory quality. The trajectory $\tau_k$ with the highest score is chosen as the predicted result in inference.

\subsection{Coarse-to-Fine Trajectory selection}
\modelName proposes a coarse-to-fine trajectory selection paradigm comprising coarse filtering and fine-grained scoring, improving model capability in distinguishing hard negative trajectories. As shown in Fig~\ref{fig:arch}, in the coarse filtering stage, the model selects several trajectory candidates based on predicted scores, similar to classic selection-based approaches. The fine-grained scoring stage then produces more accurate scores for the filtered trajectories.

\paragraph{Coarse filtering}
The coarse filtering stage scores all trajectories in the vocabulary and filters a smaller set of candidates for the next stage. Here we apply the same strategy as the previous selection-based method~\cite{li2024hydra}. The trajectory feature cross-attends with the image feature to extract planning-related information, then several prediction heads are applied on the refined trajectory feature to regress the normalized $l_2$ distance to the ground-truth human trajectory and the rule-based metric scores:
\begin{gather}
    \mathcal{E}_{\mathrm{img}} = \mathrm{Enc_{i}}(I), f_j = \mathrm{Enc_{t}}(\tau_j),\\
    g_j = \mathrm{TransDec}(\mathcal{E}_{\mathrm{img}}, f_j) \\
    s_j^{(m)} = \mathrm{Sigmoid}\left(\mathrm{head^{(m)}}(g_j)\right),
\end{gather}
where $\mathrm{Enc_{i}}$ and $\mathrm{Enc_{t}}$ are image encoder and trajectory encoder, $I$ and $\mathcal{E}_{\mathrm{img}}$ are the input image and image feature, $\tau_j$ and $f_j$ denote the trajectory and the encoded trajectory feature, $\mathrm{TransDec}$ denotes the Trajectory Decoder, which is a Transformer decoder~\cite{vaswani2017attention}, $g_j$ denotes the refined trajectory feature, $\mathrm{head^{(m)}}$ denotes the prediction head of evalutation metric $m$, $s_j^{(m)}$ denotes the prediction score of trajectory $\tau_j$ on metric $m$.

At the end of the coarse filtering stage, each trajectory $\tau_j$ corresponds to a score $s_j$ revealing its quality on end-to-end planning. We select the trajectories with top-k scores as the filtered trajectories $T_{\mathrm{filter}}= \{\tau_j \mid j \in \mathcal{I}_{\text{top}k}\}$,
where $\mathcal{I}_{\text{top}k} = \operatorname{argsort}_k(\{s_j\}_{j=1}^N)$, and the refined features $G_{\mathrm{filter}} = \{g_j \mid j \in \mathcal{I}_{\text{top}k}\}$ are utilized for fine-grained fitting.

\begin{figure*}[!t] \centering
    \includegraphics[width=0.99\textwidth]{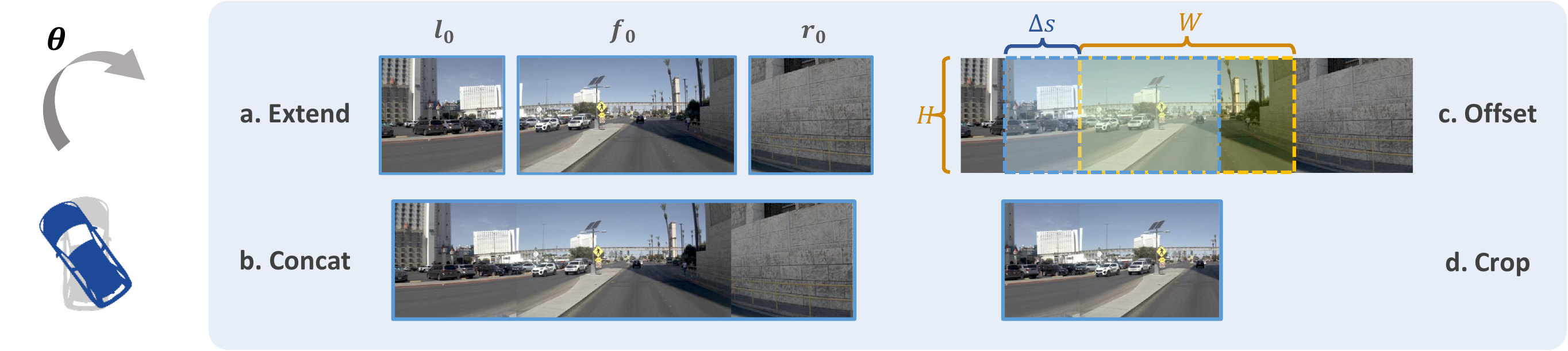}
    \caption{Rotation demonstration for input image in a 1-camera FOV setting. Our rotation-based data augmentation method generates the rotated image through horizontal shifting and cropping.}
    \label{fig:rotation}
\end{figure*}

\paragraph{Fine-grained scoring}
In this stage, a Transformer decoder similar to the first stage is applied to make fine-grained scoring and further distinguish the trajectories in the first-stage filtered candidates, which contain a large proportion of hard negatives. Specifically, we obtain the refined score from the $l$-th decoder layer output feature, and optimize with the ground truth trajectory score:
\begin{align}
    \left\{h_{j, l}\right\}_{l=1}^{n_{\mathrm{ref}}} &= \mathrm{RefineDec}\left( \mathcal{E}_{\mathrm{img}}, g_j \right)\\
    s_{j, l}^{(m)} &= \mathrm{Sigmoid}\left(\mathrm{head^{(m)}}\left(h_{j, l}\right)\right)
\end{align}
where $\mathrm{RefineDec}$ is a Transformer decoder with $n_{\mathrm{ref}}$ layers, $h_{j, l}$ is the refined feature of $\tau_j$ from the $l$-th layer of the Refinement Transformer, $s_{j, l}^{(m)}$ denotes the score of $\tau_j$ on metric $m$ output by the $l$-th decoder layer. The output trajectory $\tau_k$ is selected based on the highest last-layer output score $s_{k, n_{\mathrm{ref}}}^{(m)}$. 

The loss for the coarse-to-fine module on the original dataset is represented as:
\begin{equation}
    L_{\mathrm{ori}} = L_{\mathrm{coarse}} + L_{\mathrm{refine}},
\end{equation}
where $L_{\mathrm{coarse}}$ and $L_{\mathrm{refine}}$ respectively train the coarse filtering stage and the fine-grained scoring stage.
$L_{\mathrm{coarse}}$ consists of an imitation loss $L_{\mathrm{imi}}$~\cite{li2024hydra, li2024hydramdp_pp} and a binary cross-entropy between the predicted trajectory score and the ground truth metric score:
\begin{equation}
    L_{\mathrm{coarse}} = L_{\mathrm{imi}} + \sum_{m, i}\mathrm{BCE}(s_{i}^{(m)}, y_{i}^{(m)}),
\end{equation}
where $y_{i}^{(m)}$ and $s_{i}^{(m)}$ are the ground-truth metric score and the predicted score of trajectory $\tau_i$ on metric $m$.
$L_{\mathrm{refine}}$ is similar to $L_{\mathrm{coarse}}$, while it only considers the filtered trajectories $T_{\mathrm{filter}}$ rather than the entire vocabulary, and is applied on each decoder layer output.

\subsection{Rotation-based data augmentation}

To mitigate the data imbalance, we introduce an end-to-end rotation-based augmentation pipeline, where a 2D horizontal view transformation is applied to the sensor input data to simulate the ego-vehicle rotation in the 3D space. This approach easily synthesizes more challenging scenarios and diversifies the driving scenarios, enabling the model to precisely select trajectories regardless of the vehicle orientation.

As shown in Fig~\ref{fig:rotation}, for each scenario, we sample a random angle $\theta$ from a uniform distribution $U[-\Theta, \Theta]$, where $\Theta$ is the angle boundary. The positive angle indicates the leftward rotation of the ego vehicle. Camera images corresponding to the original field of view (FOV) along with images from two extended views are concatenated to simulate a ``pseudo panoramic view'', then the input image is cropped from the concatenated image according to a shifting window based on $\theta$.
For example, for a 1-camera FOV input setting, we choose these three cameras as our input: $f$ (front), $l_0$ (front-left), and $r_0$ (front-right), where $f$ corresponds to the original field of view, and $l_0$ and $r_0$ correspond to the extended view.

In the synthesized rotated scenario, the ground truth human trajectory $(u_1, u_2, \dots, u_l)$  is generated by a trivial rotation approach: under a specific rotation angle $\theta$, each location $u_j$ of the human trajectory $T_h$ applies a 2D-rotation transformation surrounding the initial vehicle position $u_0$, and the rotation angle is $-\theta$, ensuring the world coordinate of the trajectory remains unchanged.
Given augmented camera views and the prediction of our model, we calculate a loss $L_{\mathrm{aug}}$ for training, which has the same formulation as $L_{\mathrm{ori}}$.

\subsection{Self-distillation with Soft-labeling}

Instead of training the selection-based model to fit a hard decision boundary of trajectory evaluation, which can impair training stability, we propose a self-distillation framework with teacher-generated soft labels to stabilize model training. The self-distillation framework consists of a teacher and a student model, both sharing the same architecture. The student is updated via standard gradient descent, whereas the teacher is updated using an exponential moving average (EMA) of the student’s parameters. 

During training, the student receives noisier input, including both original and augmented data to calculate $L_{\mathrm{ori}}$ and $L_{\mathrm{aug}}$, as shown in Fig~\ref{fig:arch}. The teacher only receives original data to generate scores served as soft labels, where a clipping threshold $\delta_m$ is introduced to control the gap between the teacher output and the ground-truth:
\begin{gather}
    \hat{y}_{i}^{(m)} = y_{i}^{(m)} + \mathrm{clip}\left( s_{i, \mathrm{teacher}}^{(m)} - y_{i}^{(m)}, -\delta_m, \delta_m \right) \label{equ:soft-label} \\
    L_{\mathrm{soft}} = L_{\mathrm{imi-soft}} +\sum_{m, i}{\mathrm{BCE}(s_{i}^{(m)}, \hat{y}_{i}^{(m)})},
\end{gather}
where $s_{i, \mathrm{teacher}}^{(m)}$ is the score of trajectory $i$ on metric $m$ predicted by the teacher, $y_{i}^{(m)}$ is the ground-truth score of trajectory $i$ on metric $m$, $\mathrm{clip}\left(\cdot,\alpha, \beta\right)$ denotes clipping the input value to the interval $[\alpha, \beta]$, and $\hat{y}_{i}^{(m)}$ is the soft label. $L_{\mathrm{imi-soft}}$ is a soft version of $L_{\mathrm{imi}}$, where the human trajectory is shifted toward the teacher model's output trajectory for at most 1 meter, and is adopted to calculate the imitation loss. 
The overall training loss of the student model is:
\begin{equation}
    L = L_{\mathrm{ori}} + L_{\mathrm{aug}} +  L_{\mathrm{soft}}
\end{equation}

During inference, the teacher model is utilized to output the planning trajectories.

\begin{table*}[!t]
    
    \centering
    \small
    \begin{tabular}{c|c|*{6}{c}}
        \toprule
        Method & Backbone & NC $\uparrow$ & DAC $\uparrow$ & EP $\uparrow$ & TTC $\uparrow$ & C $\uparrow$ & PDMS $\uparrow$ \\
        \midrule
        Human Agent & — & 100 & 100 & 87.5 & 100 & 99.9 & 94.8 \\
        \midrule
        Transfuser~\shortcite{chitta2022transfuser} & ResNet34 & 97.7 & 92.8 & 79.2 & 92.8 & 100 & 84.0 \\
        UniAD~\shortcite{hu2023planning} & ResNet34& 97.8 & 91.9 & 78.8 & 92.9 & 100 & 83.4 \\
        VADv2~\shortcite{chen2024vadv2} & ResNet34 & 97.9 & 91.7 & 77.6 & 92.9 & 100 & 83.0 \\
        LAW~\shortcite{li2024enhancing} & ResNet34 & 96.4 & 95.4 & 81.7 & 88.7 & 99.9 & 84.6 \\
        DRAMA~\shortcite{yuan2024drama} & ResNet34 & 98.0 & 93.1 & 80.1 & 94.8 & 100 & 85.5 \\
        Hydra-MDP~\shortcite{li2024hydra} & ResNet34 & 98.3 & 96.0 & 78.7 & 94.6 & 100 & 86.5 \\
        DiffusionDrive~\shortcite{liao2024diffusiondrive} & ResNet34 & 98.2 & 96.2 & 82.2 & 94.7 & 100 & 88.1 \\
        \textbf{\modelName} & ResNet34 & 97.8 & 97.3 & 86.7 & 93.6 & 100 & \textbf{89.9 (+1.8)} \\
        \midrule
        Hydra-MDP~\shortcite{li2024hydra} & V2-99 & 98.4 & 97.8 & 86.5 & 93.9 & 100 & 90.3 \\
        \textbf{\modelName} & V2-99 & 98.0 & 98.2 & 90.0 & 94.2 & 100 & \textbf{92.1 (+1.9)} \\
        \midrule
        Hydra-MDP~\shortcite{li2024hydra} & ViT-L & 98.4 & 97.7 & 85.0 & 94.5 & 100 & 89.9 \\
        \textbf{\modelName} & ViT-L & 98.6 & 98.6 & 91.3 & 95.5 & 100 & \textbf{93.5 (+3.6)} \\
        \bottomrule
    \end{tabular}
    \caption{Evaluation on NAVSIM v1. Results are grouped by backbone types.}
    \label{tab:navsim_v1}
\end{table*}

\begin{table*}[!t]
    \centering
    \small
    \begin{tabular}{c|c|*{9}{c}|cc}
        \toprule
        Method & Backbone & NC $\uparrow$ & DAC $\uparrow$ & DDC $\uparrow$ & TL $\uparrow$ & EP $\uparrow$ & TTC $\uparrow$ & LK $\uparrow$ & HC $\uparrow$ & EC $\uparrow$ & EPDMS $\uparrow$ \\
        \midrule
        Human Agent & — & 100 & 100 & 99.8 & 100 & 87.4 & 100 & 100 & 98.1 & 90.1 & 90.3 \\
        \midrule
        Ego Status MLP & — & 93.1 & 77.9 & 92.7 & 99.6 & 86.0 & 91.5 & 89.4 & 98.3 & 85.4 & 64.0 \\
        \midrule
        Transfuser~\shortcite{chitta2022transfuser} & ResNet34 & 96.9 & 89.9 & 97.8 & 99.7 & 87.1 & 95.4 & 92.7 & 98.3 & 87.2 & 76.7 \\
        HydraMDP++~\shortcite{li2024hydramdp_pp} & ResNet34 & 97.2 & 97.5 & 99.4 & 99.6 & 83.1 & 96.5 & 94.4 & 98.2 & 70.9 & 81.4 \\
        \textbf{\modelName} & ResNet34 & 97.5 & 96.5 & 99.4 & 99.6 & 88.4 & 96.6 & 95.5 & 98.3 & 77.0 & \textbf{83.1 (+1.7)} \\
        \midrule
        HydraMDP++~\shortcite{li2024hydramdp_pp} & V2-99 & 98.4 & 98.0 & 99.4 & 99.8 & 87.5 & 97.7 & 95.3 & 98.3 & 77.4 & 85.1 \\
        \textbf{\modelName} & V2-99 & 97.8 & 97.9 & 99.5 & 99.9 & 90.6 & 97.1 & 96.6 & 98.3 & 77.9 & \textbf{86.0 (+0.9)} \\
        \midrule
        HydraMDP++~\shortcite{li2024hydramdp_pp} & ViT-L & 98.5 & 98.5 & 99.5 & 99.7 & 87.4 & 97.9 & 95.8 & 98.2 & 75.7 & 85.6 \\
        \textbf{\modelName} & ViT-L & 98.4 & 98.6 & 99.6 & 99.8 & 90.5 & 97.8 & 97.0 & 98.3 & 78.6 & \textbf{87.1 (+1.5)} \\
        \bottomrule
    \end{tabular}
    \caption{Evaluation on NAVSIM v2. Results are grouped by backbone types.}
    \label{tab:navsim_v2}
\end{table*}

\begin{table}[ht]
    \small
    \selectfont
    \centering
    \begin{tabular}{c|cccc}
    \toprule
    Method & DS $\uparrow$ & SR $\uparrow$ & Eff. $\uparrow$ & Comf. $\uparrow$ \\ 
    \midrule
    DriveAdapter~\shortcite{jia2023driveadapter}  & 64.22 & 33.08 & 70.22 & 16.01 \\ 
    Hydra-NeXt~\shortcite{li2025hydranext}  & 73.86 & 50.00 & 197.76 & 20.68 \\ 
    Orion~\shortcite{fu2025orion}  & 77.74 & 54.62 & 151.48 & 17.38 \\ 
    AutoVLA~\shortcite{zhou2025autovla} & 78.84 & 57.73 & 146.93 & \textbf{39.33} \\
    \textbf{DriveSuprim} & \textbf{83.02} & \textbf{60.00} & \textbf{238.78} & 20.89 \\ 
    \bottomrule
    \end{tabular}
    \caption{Evaluation on Bench2Drive. 
    }
    \label{tab:b2d}
\end{table}

\section{Experiments}

In this section, we first introduce our implementation details. Next, we show the superior performance of \modelName on NAVSIM and Bench2Drive, and ablation studies on NAVSIM are listed to validate the effectiveness of the proposed modules. Finally, we produce some visualization results to intuitively show the advantages of our method.

\subsection{Implementation Details}

\paragraph{Dataset and Metrics}
We conduct experiments mainly on the NAVSIM~\cite{dauner2024navsim} and Bench2Drive~\cite{jia2024bench} benchmark.
NAVSIM includes two evaluation metrics, leading to NAVSIM v1 and NAVSIM v2. The evaluation metric of NAVSIM v1 is the PDM Score (PDMS). Each predicted trajectory is sent to a simulator to collect different rule-based metrics, which are weighted aggregated to get the final PDMS.
NAVSIM v1 includes 5 metrics: no collisions (NC), drivable area compliance (DAC), ego progress (EP), time-to-collision (TTC), and comfort (C). NAVSIM v2 further introduces 4 extended metrics, including driving direction compliance (DDC), traffic light compliance (TLC), lane keeping (LK) and extended comfort (EC).
Aggregating all these subscores leads to the EPDMS metric.

Bench2Drive is a closed-loop benchmark evaluating 220 short routes in CARLA v2. The evaluation metrics include Success Rate, Driving Score, Efficiency and Comfortness. 
More details about datasets and metrics are in Appendix A.

\paragraph{Model Details and Training Details}
We conduct our methods on three different backbones as the image encoder $\mathrm{Enc}_i$, including ResNet34~\cite{he2016deep}, VoVNet~\cite{lee2019an}, and ViT-Large~\cite{dosovitskiy2021an}. 
A 2-layer MLP is leveraged as $\mathrm{Enc}_t$ to encode each trajectory in the vocabulary. The trajectory decoder $\mathrm{TransDec}$ and refinement decoder $\mathrm{RefineDec}$ are both 3-layer Transformer Decoders with 256 hidden dimensions. 
The MLP prediction head $\mathrm{head}^{(m)}$ predicts the normalized $l_2$ distance to the human ground-truth trajectory, and each subscore in NAVSIM except for EC. 
The number of trajectories in the vocabulary and filtered trajectories of the coarse filtering stage are set to 8192 and 256. We adopt a 3-camera FOV setting, with images from $l_0$, $f$, and $r_0$ cameras as the input. The rotation angle boundary $\Theta$ is set to $\pi / 6$ in the augmentation pipeline. The threshold $\delta_m$ in soft-labeling is set to 0.15.

We train our model on 8 NVIDIA A100. We use Adam for model training, the batch size on a GPU is 8, and the learning rate is set to $7.5\times 10^{-5}$. More model details and training details are shown in Appendix B.

\subsection{Main Results}

\paragraph{Result on NAVSIM}
Tab~\ref{tab:navsim_v1} shows the performance of \modelName on the NAVSIM benchmark. our method reaches 89.9\% PDMS with the ResNet34 backbone, surpassing DiffusionDrive by 1.8\%. Moreover, \modelName with a stronger Vit-Large backbone can reach 93.5\% PDMS.
Results in Table~\ref{tab:navsim_v2} show that our model can also reach the SOTA result on the more challenging NAVSIM v2 benchmark. On the EPDMS metric, \modelName surpasses previous SOTA methods by 1.7\%, 0.9\%, and 1.5\%, respectively. 
\modelName reaches 27.2 FPS and 12.5 FPS with ResNet34 and ViT-Large backbone, achieving real-time planning capability.

\paragraph{Result on Bench2Drive}
Evaluation on Bench2Drive shows the close loop planning capability of our method. Based on the CARLA-Garage dataset~\cite{jaeger2023hidden, sima2024drivelm} and the TF++ framework~\cite{jaeger2023hidden}, our approach adopts the two-stage trajectory prediction for longitudinal control. Tab~\ref{tab:b2d} shows that \modelName achieves 83.02 and 60.00 scores on Driving Score and Success Rate, with 238.78 Efficiency and 20.89 Comfortness, outperforming the previous SOTA significantly.

\begin{table*}[!t]\centering
    \small
    \begin{tabular}{ccc|*{9}{c}|c}
        \toprule
        \makecell{Multi-\\stage} & \makecell{Aug \\ Data} & \makecell{Self-\\distill} & NC $\uparrow$ & DAC $\uparrow$ & DDC $\uparrow$ & TL $\uparrow$ & EP $\uparrow$ & TTC $\uparrow$ & LK $\uparrow$ & HC $\uparrow$ & EC $\uparrow$ & EPDMS $\uparrow$ \\
        \midrule
        \CrossM & \CrossM & \CrossM & 97.2 & 97.5 & 99.4 & 99.6 & 83.1 & 96.5 & 94.4 & 98.2 & 70.9 & 81.4  \\
        \CM & \CrossM & \CrossM & 97.5 & 96.4 & 99.1 & 99.6 & 87.0 & 96.4 & 95.3 & 98.2 &  75.0 & 82.4  \\
        \CM & \CM & \CrossM & 96.9 & 96.9 & 99.4 & 99.6 & 87.9 & 96.0 & 95.5 & 98.3 & 76.5 & 82.7 \\
        \CM & \CM & \CM & 97.5 & 96.5 & 99.4 & 99.6 & 88.4 & 96.6 & 95.5 & 98.3 & 77.0 & 83.1 \\
        \bottomrule
    \end{tabular}
    \caption{Ablation study on different proposed modules. ``Multi-stage'' denotes using coarse-to-fine selection, ``Aug Data'' denotes introducing rotation-based augmentation data, and ``Self-distill'' denotes adopting self-distillation.
    }
    \label{tab:ablation-module}
\end{table*}

\begin{table}[!t]\centering
    \small
    \begin{tabular}{l|c}
        \toprule
        & EPDMS $\uparrow$ \\
        \midrule
        Single-stage & 81.4 \\
        + 6 layer decoder & 81.7 \\
        + Layer-wise scoring & 81.9 \\
        + Trajectory filtering & 82.4 \\
        \bottomrule
    \end{tabular}
    \caption{Ablation study of the evolution from single-stage selection to coarse-to-fine selection.
    }
    \label{tab:ablation-evolution}
\end{table}

\subsection{Ablation Studies}

\paragraph{Ablation on different modules}
We conduct ablation studies on the modules and approaches we propose, and the result is shown in Tab~\ref{tab:ablation-module}. Compared to baseline, adopting multi-stage refinement leads to 1.0\% performance gain on PDMS, and the improvements further gained by augmented data and the self-distillation framework are 0.3\% and 0.4\%.

\paragraph{Effectiveness of coarse-to-fine selection}
We evolve our model from conventional single-stage selection to coarse-to-fine selection, to eliminate the effect of parameter number increase and validate the effectiveness of the coarse-to-fine trajectory selection paradigm. Tab~\ref{tab:ablation-evolution} reveals that increasing the number of decoder layers from 3 to 6 brings only 0.3\% improvement in EPDMS, while introducing layer-wise scoring and trajectory filtering leads to a more substantial 0.7\% gain, proving that the performance boost comes from the coarse-to-fine mechanism rather than model size increase.

\begin{figure}[!b]
    \centering
    \includegraphics[width=0.46\textwidth]{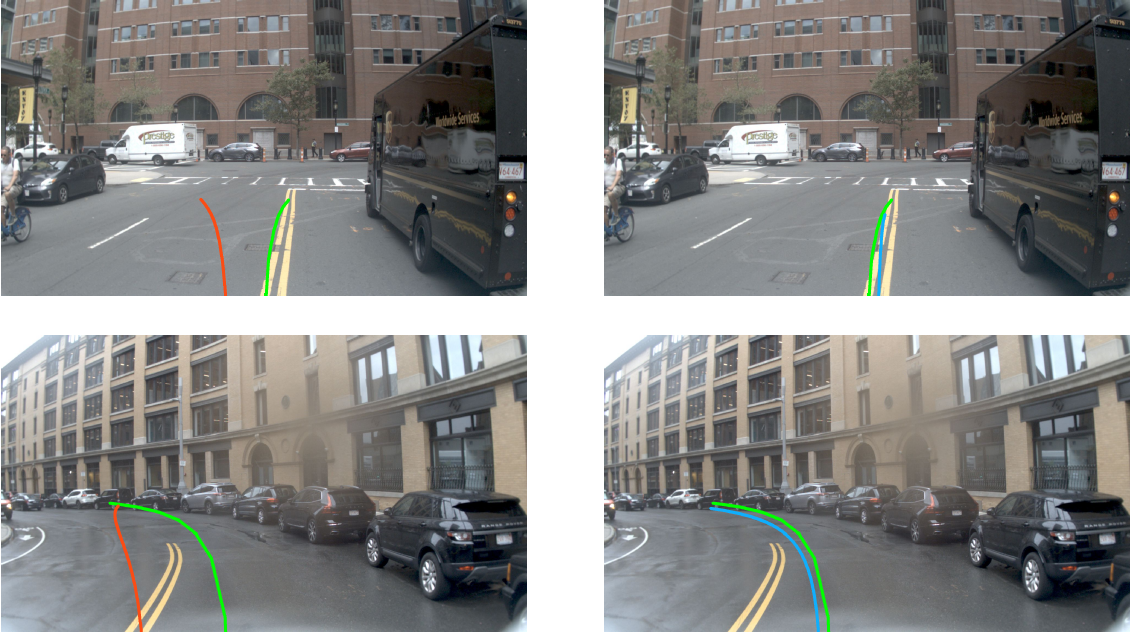}
    \caption{Visualization results across various challenging scenarios. 
    In each example, the green trajectory represents the ground truth from the human expert, the red trajectory is generated by Hydra-MDP++, and the blue trajectory is produced by \modelName.}
    \label{fig:vis}
\end{figure}

\paragraph{Ablation on refinement settings and soft label threshold}
We further conduct comprehensive ablation studies on the refinement approach and soft-labeling.
Utilizing a 3-layer refinement Transformer decoder and filtering 256 trajectories for the refinement stage leads to the best result.
For the teacher soft label, choosing 0.15 as the threshold leads to the best EPDM score. Results are shown in Appendix C.

\begin{table}[!t]
    \centering
    \small
    \setlength{\tabcolsep}{1mm}
    \begin{tabular}{c|ccc}
        \toprule
        Method & $\mathrm{NAVTEST}_{\mathrm{l}}$ & $\mathrm{NAVTEST}_{\mathrm{f}}$ & $\mathrm{NAVTEST}_{\mathrm{r}}$ \\
        \midrule
        Hydra-MDP++ & 68.7 & 87.2 & 77.7 \\
        \modelName & \textbf{71.6 (+2.9)} & 88.1 (+0.9) & \textbf{79.7 (+2.0)} \\
        \bottomrule
    \end{tabular}
    \caption{EPDMS on three dataset splits. $\mathrm{NAVTEST}_{\mathrm{l}}$, $\mathrm{NAVTEST}_{\mathrm{f}}$, and $\mathrm{NAVTEST}_{\mathrm{r}}$ involve left-turning scenarios, near-forward scenarios, and right-turning scenarios.
    }
    \label{tab:split}
\end{table}

\subsection{Visualization and Analysis}

\paragraph{Visualization Results}
Fig~\ref{fig:vis} shows qualitative visualization results on NAVSIM comparing \modelName with the selection-based approach Hydra-MDP++. The images in the first and second columns are the results of Hydra-MDP++ and \modelName.
The first row illustrates a challenging overtaking scenario near a crossroad, where \modelName correctly overtakes while Hydra-MDP++ mistakenly turns left and risks collision. In the second row, \modelName outperforms in a sharp turn with smooth and accurate trajectories. These results demonstrate \modelName not only performs well in challenging scenarios by choosing precise trajectories, but also excels in handling sharp turns with high accuracy. More visualization results are shown in Appendix D.

\paragraph{Superior performance on turning scenarios}
Tab.~\ref{tab:split} demonstrates the superior performance of \modelName in turning scenarios. We divide the test dataset into three subsets based on the turning angle of the ground-truth trajectories: $\mathrm{NAVTEST}_{\mathrm{l}}$ (left turns exceeding 30 degrees), $\mathrm{NAVTEST}_{\mathrm{f}}$ (near-forward trajectories), and $\mathrm{NAVTEST}_{\mathrm{r}}$ (right turns exceeding 30 degrees). Performance improvements of \modelName are more pronounced in turning scenarios than in near-straight ones, highlighting enhanced ability to handle turning maneuvers.

\paragraph{Trajectory distribution comparison}
We illustrate the trajectory frequency distribution in Appendix D. Results show that in the original dataset, trajectories are predominantly concentrated in the forward or near-forward direction, while in the augmented dataset, trajectories across all directions appear with similar frequency.

\section{Conclusion}
\label{sec:conclusion}
We present \modelName, a novel framework for end-to-end planning. By introducing coarse-to-fine selection and an integrated training pipeline with rotation-based data augmentation and soft-label self-distillation, \modelName significantly enhances the model's ability to distinguish hard negatives and precisely select trajectories, and performs well in scenarios involving sharp turns. Extensive experiments on the NAVSIM and Bench2Drive benchmark demonstrate that our approach outperforms prior methods by a substantial margin.

\section{Appendix}

\subsection{A. More information about Datasets and Metrics}
\paragraph{NAVSIM}
NAVSIM~\cite{dauner2024navsim} is split into navtrain and navtest, containing 103k and 12k samples for model training and testing.
There are two different evaluation metrics on NAVSIM, leading to NAVSIM v1 and NAVSIM v2. The evaluation metric of NAVSIM v1 is the PDM Score (PDMS). Each predicted trajectory is sent to a simulator to collect different rule-based subscores, which are multiplied or weighted aggregated to get the final PDMS:
\begin{equation}
    \mathrm{PDMS} = \left( \prod_{m \in S_{P}}{\mathrm{s}_m} \right)  \times \left( \frac{ \sum_{w \in S_{A}}{\mathrm{w}_w \times \mathrm{s}_w} }{ \sum_{w \in S_{A}}{\mathrm{w}_w} } \right),
\end{equation}
where $S_{P}$ and $S_{A}$ denote the multiplier subscore set and the weighted average subscore set, $\mathrm{s}$ denotes the subscore, and $\mathrm{w}$ is the weight of each subscore. In NAVSIM v1, $S_{P}$ comprises two subscores: no collisions (NC) and drivable area compliance (DAC), and $S_{A}$ comprises ego progress (EP), time-to-collision (TTC), and comfort (C).
NAVSIM v2 introduces 4 more subscores, including driving direction compliance (DDC) and traffic light compliance (TLC) in $S_{P}$, and lane keeping (LK) and extended comfort (EC) in $S_{A}$. The comfort subscore is also revised to become the history comfort (HC). Aggregating all the above subscores leads to the $\mathrm{EPDMS}$ metric.

\paragraph{Bench2Drive}
Bench2Drive~\cite{jia2024bench2drive} is a recently proposed closed-loop evaluation benchmark for end-to-end autonomous driving, which includes 220 short routes spanning 44 interactive driving scenarios. Each route is around 150 meters. This short-distance route design allows for a comprehensive and accurate analysis of the planning model's capability. The evaluation metrics include the Success Rate (SR), Driving Score (DS), Efficiency, and Comfortness. The Success Rate measures the proportion of routes executed correctly by the planner, the Driving Score considers the planner's performance from route completion and rule compliance, the Efficiency checks whether the vehicle speed is too slow, and the Comfortness measures the smoothness by comparing the trajectory with human behaviors.

\subsection{B. Supplementary Implementation Details}
\label{sec:supp_training_details}

\paragraph{Model details}
We conduct our methods on three different backbones as the image encoder $\mathrm{Enc}_i$, including ResNet34~\cite{he2016deep}, VoVNet~\cite{lee2019an}, and ViT-Large~\cite{dosovitskiy2021an}. 
The input images are resized to a resolution of $2048\times 512$ for ResNet34 and VoVNet, and $1024\times 256$ for Vit-Large. The ViT-Large backbone is pre-trained through Depth Anything~\cite{yang2024depth}.

\paragraph{Training and inference pipelines}
We adopt two different training pipelines for different backbones. For the models loading the pre-trained backbones, including VoVNet and Vit-Large, we train our model for 6 epochs. The EMA momentum $m$ is increased linearly from 0.992 to 0.996 in the first 3 epochs, and then fixed to 0.998. The model with the ResNet-34 backbone trained from scratch is trained for 10 epochs. The EMA update is applied after the training for 3 epochs, which means that $m$ is set to 0 in the first 3 epochs, then increased from 0.992 to 0.996 in the subsequent 3 epochs, then fixed to 0.998.

During inference, we compute a final score for each trajectory by linearly combining the predicted scores across different metrics, and the filtered and final trajectories are selected based on the final score. Specifically, for the imitation learning metric and the multiplier metrics introduced in NAVSIM~\cite{dauner2024navsim}, we apply a coefficient to the logarithm of the predicted score. For the weighted average metric in NAVSIM, the coefficient is applied to the predicted score, followed by an additional logarithmic process of the sum. The overall inference score can be formulated as:
\begin{equation}
    s_i = \sum_{m}{\lambda^{(m)}\log {s_i^{(m)}}} +\lambda_{\mathrm{avg}}\log\left({\sum_{n}{\lambda^{(n)}s_i^{(n)}}}\right),
\end{equation}
where $m$ denotes the imitation metric and the multiplier metric, $n$ denotes the weighted average metric, $\lambda^{(m)}$ and $\lambda^{(n)}$ denote the coefficient, and $s_i$ denotes the final combined prediction score. $\lambda_{\mathrm{avg}}$ is set to 8.0 and 6.0 in NAVSIM v1 and v2, respectively. The detailed coefficients used during inference on NAVSIM v1 and v2 are shown in Tab~\ref{tab:coefficient-v1} and Tab~\ref{tab:coefficient-v2}.

\begin{table}[!t]
    \centering
    \small
    \begin{tabular}{c|*{6}{c}}
        \toprule
        & \multirow{2}{*}{Imi} & \multicolumn{2}{c}{Mul} & \multicolumn{3}{c}{Avg}  \\
        \cmidrule[0.5pt](rl){3-4}
        \cmidrule[0.5pt](rl){5-7}
        & & NC & DAC & EP & TTC & C \\
        \midrule
        coefficient & 0.05 & 0.5 & 0.5 & 5.0 & 5.0 & 2.0 \\
        \bottomrule
    \end{tabular}
    \caption{The inference coefficients on each metric of NAVSIM v1. ``Imi'' denotes the imitation metric, ``Mul'' denotes the  multiplied penalties, and ``Avg'' denotes the weighted averages.}
    \label{tab:coefficient-v1}
\end{table}

\begin{table}[!t]
    \centering
    \small
    \setlength{\tabcolsep}{1.2mm}
    \begin{tabular}{c|*{10}{c}}
        \toprule
        & \multirow{2}{*}{Imi} & \multicolumn{4}{c}{Mul} & \multicolumn{4}{c}{Avg}  \\
        \cmidrule[0.5pt](rl){3-6}
        \cmidrule[0.5pt](rl){7-10}
        & & NC & DAC & DDC & TL & EP & TTC & LK & HC \\
        \midrule
        coefficient & 0.02 & 0.5 & 0.5 & 0.3 & 0.1 & 5.0 & 5.0 & 2.0 & 1.0 \\
        \bottomrule
    \end{tabular}
    \caption{The inference coefficients on each metric of NAVSIM v2. ``Imi'' denotes the imitation metric, ``Mul'' denotes the  multiplied penalties, and ``Avg'' denotes the weighted averages.}
    \label{tab:coefficient-v2}
\end{table}

\subsection{C. Supplementary Experiment Result}

\paragraph{Ablation on refinement and soft label settings}
We further conduct comprehensive ablation studies on the refinement approach and soft-labeling. Tab~\ref{tab:ablation-stage_layer-top_k} shows that utilizing a 3-layer refinement Transformer decoder and filtering 256 trajectories for the refinement stage leads to the best result.
Tab~\ref{tab:ablation-threshold} and Tab~\ref{tab:ablation-imi_shift} show ablations on soft label. For the teacher soft label, choosing 0.15 as the clipping threshold and adopting a 1-meter shift in soft imitation learning lead to the best EPDM score.

\begin{table}[!t]
    \centering
    \small
    \begin{tabular}{cc|c}
        \toprule
        \makecell{Stage \\ Layer} & \makecell{Top-K} & EPDMS $\uparrow$ \\
        \midrule
        1 & 256 & 83.0 \\
        3 & 256 & \textbf{83.1} \\
        6 & 256 & 82.6 \\
        3 & 64 & 81.8 \\
        3 & 512 & 82.9 \\
        3 & 1024 & 83.0 \\
        \bottomrule
    \end{tabular}
    \caption{Ablation on refinement setting. ``Stage Layer'' is the layer number of the Refinement Decoder, and ``Top-K'' denotes the number of trajectories selected by the coarse filtering stage.}
    \label{tab:ablation-stage_layer-top_k}
\end{table}

\begin{table}[!t]
    \centering
    \small
    \begin{tabular}{c|c}
        \toprule
        $\delta_m$ & EPDMS $\uparrow$ \\
        \midrule
        0.00 & 82.6 \\
        0.15 & \textbf{83.1} \\
        0.30 & 83.0 \\
        0.70 & 82.9 \\
        1.00 & 82.8 \\
        \bottomrule
    \end{tabular}
    \caption{Results with different soft label thresholds. $\delta_m$ denotes the soft label threshold in the Equation~\ref{equ:soft-label}.}
    \label{tab:ablation-threshold}
\end{table}

\begin{table}[!t]
    \centering
    \small
    \begin{tabular}{c|c}
        \toprule
        $d_i$ & EPDMS $\uparrow$ \\
        \midrule
        0.0 & 82.6 \\
        0.5 & 83.0 \\
        1.0 & \textbf{83.1} \\
        2.0 & 82.7 \\
        \bottomrule
    \end{tabular}
    \caption{Results with different soft imitation trajectory shift. $d_i$ denotes the soft imitation trajectory shift in $L_\mathrm{imi-soft}$ in the Equation 10.}
    \label{tab:ablation-imi_shift}
\end{table}

\begin{figure*}[!t]
    \centering
    \includegraphics[width=0.86\textwidth]{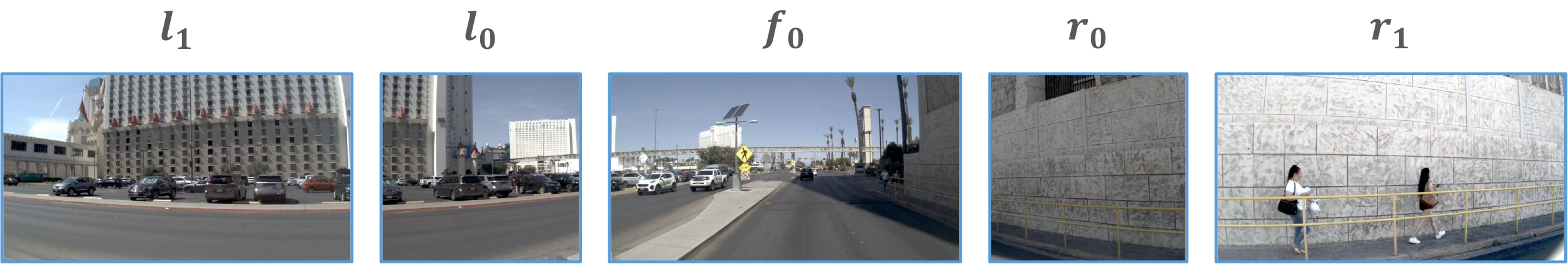}
    \caption{Images from the five cameras.}
    \label{fig:fovs}
\end{figure*}

\paragraph{Model performance with different FOV settings}
We show the model performance with different FOV settings in Tab~\ref{tab:ablation-fov}. The FOV is revealed by the number of cameras. As shown in Fig~\ref{fig:fovs}, five cameras are involved in these settings: $f$ (front camera), $l_0$ (front-left camera), $l_1$ (left camera), $r_0$ (front-right camera), and $r_1$ (right camera). The 1-camera setting uses $f_0$ as input, the 3-camera setting uses $l_0$, $f_0$ and $r_0$ as input, while the 5-camera setting uses $l_1$, $l_0$, $f_0$, $r_0$ and $r_1$ as input. Experimental results show that the 5-camera setting leads to the best performance.

\paragraph{Superiority of self-distillation}
\modelName utilizes a self-distillation framework to mitigate the optimization challenges posed by hard binary ground-truth labels. We compare our self-distillation framework with other smoothing techniques to validate its superiority. We adopt temperature-scaled cross-entropy and vallina label smoothing, respectively. The temperature is set to 2.0, and the label smoothing coefficient is set to 0.1. As shown in Tab~\ref{tab:comparison-smoothing}, our proposed self-distillation framework achieves the best performance on EPDMS.

\paragraph{Parameter number and inference time}
We list the number of parameters and inference time of \modelName in Tab~\ref{tab:inference-speed}. Compared to Hydra-MDP++, our model introduces only a minor increase of approximately 7M parameters. Moreover, the FPS drop due to the parameter increase is acceptable. \modelName maintains the real-time planning capability.

\subsection{D. Visualization}
\begin{figure*}[!t]
    \centering
    \includegraphics[width=0.98\textwidth]{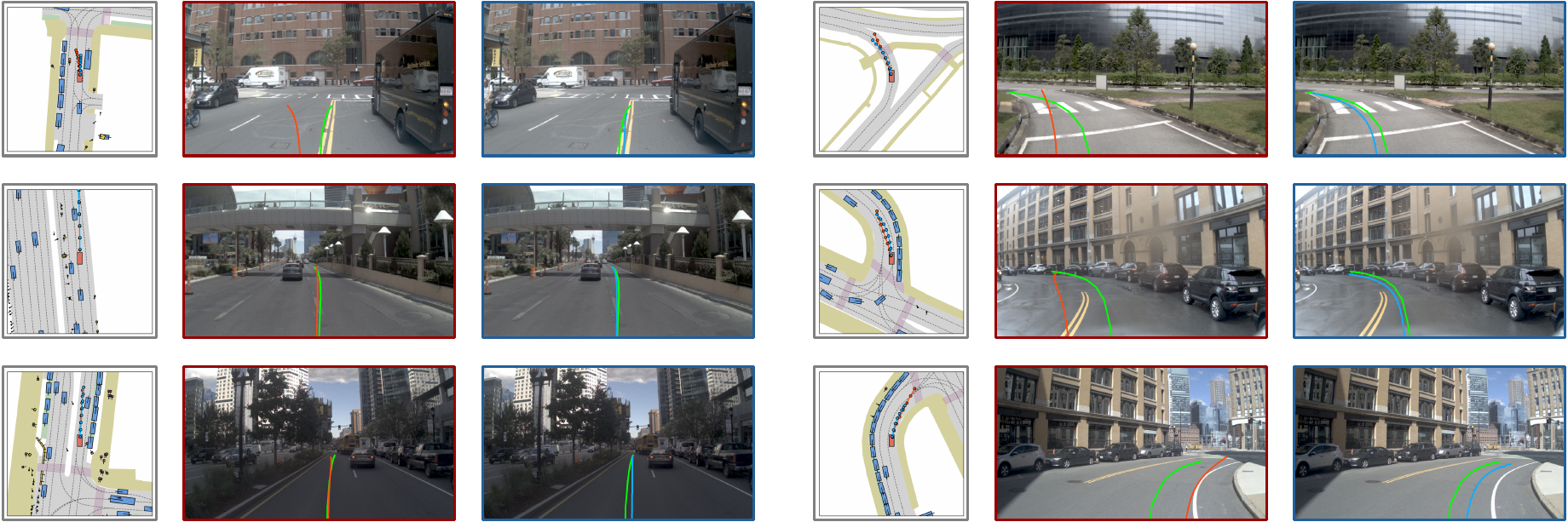}
    \caption{Visualization results across various challenging or sharp turning scenarios. In each example, the green trajectory represents the ground truth from the human expert, the red trajectory is generated by the classic selection-based method, and the blue trajectory is produced by \modelName. Zoom in for a better view.}
    \label{fig:supp_vis}
\end{figure*}

\paragraph{More visualization result}
Fig~\ref{fig:supp_vis} shows qualitative visualization results comparing our method with the selection-based approach Hydra-MDP++~\cite{li2024hydramdp_pp}. In each example, the image with a grey border shows the BEV visualization, the image with a red border is the result of Hydra-MDP++, and the image with a blue border is our result.

The first three columns on the left show challenging driving scenarios with complex vehicle interaction. In the first row, the ego vehicle attempts to overtake a leading vehicle just before a crossroad. \modelName successfully completes the overtaking maneuver, whereas the classic selection-based method incorrectly does a left turn, potentially leading to a collision. The second and third rows further highlight \modelName's ability to distinguish between trajectories that appear right at first glance, i.e., the ``hard negatives''. Although both models generate similar trajectories, the trajectories produced by Hydra-MDP++ deviate from the road centerline at the endpoint, whereas our method generates near-perfect trajectories that closely align with the human expert’s ground truths. The last three columns highlight the superior performance of our model in sharp-turning scenarios. \modelName consistently produces smooth and accurate turning trajectories even under large turning angles.

These qualitative results demonstrate that \modelName not only performs well in complex and challenging scenarios by generating precise trajectories, but also excels in handling sharp turns with high accuracy.

\begin{table}[!t]
    \centering
    \small
    \begin{tabular}{c|c}
        \toprule
        Smoothing Method & EPDMS $\uparrow$\\
        \midrule
        Label Smoothing & 82.5 \\
        Temperature-scaled CE & 82.7 \\
        Self-distillation & \textbf{83.1} \\
        \bottomrule
    \end{tabular}
    \caption{Comparison of different smoothing methods.}
    \label{tab:comparison-smoothing}
\end{table}

\begin{table}[!t]
    \centering
    \setlength{\tabcolsep}{1.5mm}
    \small
    \begin{tabular}{c|c|c|cc}
        \toprule
        Model & Backbone & EPDMS $\uparrow$ & Param. $\downarrow$ & FPS $\uparrow$ \\
        \midrule
        HydraMDP++ & ResNet34 & 81.4 & 53M & 32.0 \\
        HydraMDP++ & V2-99 & 85.1 & 103M & 24.0 \\
        HydraMDP++ & Vit-L & 85.6 & 337M & 14.4 \\
        \midrule
        \modelName & ResNet34 & 83.1 & 61M & 27.2 \\
        \modelName & V2-99 & 86.0 & 110M & 22.1 \\
        \modelName & Vit-L & 87.1 & 345M & 12.5 \\
        \bottomrule
    \end{tabular}
    \caption{Comparison of parameter and inference speed.}
    \label{tab:inference-speed}
\end{table}

\begin{table*}[!t]\centering
    \small
    \begin{tabular}{c|*{9}{c}|c}
        \toprule
        Camera Number & NC $\uparrow$ & DAC $\uparrow$ & DDC $\uparrow$ & TL $\uparrow$ & EP $\uparrow$ & TTC $\uparrow$ & LK $\uparrow$ & HC $\uparrow$ & EC $\uparrow$ & EPDMS $\uparrow$ \\
        \midrule
        1 Camera & 97.1 & 96.4 & 99.4 & 99.5 & 88.6 & 96.2 & 96.0 & 98.3 & 76.2 & 82.9 \\
        3 Cameras & 97.5 & 96.5 & 99.4 & 99.6 & 88.4 & 96.6 & 95.5 & 98.3 & 77.0 & 83.1 \\
        5 Cameras & 97.3 & 96.8 & 99.4 & 99.5 & 88.3 & 96.4 & 96.0 & 98.3 & 76.6 & \textbf{83.2} \\
        \bottomrule
    \end{tabular}
    \caption{Results with different FOVs (number of cameras).}
    \label{tab:ablation-fov}
\end{table*}

\begin{figure}[!b]
    \centering
    \includegraphics[width=0.47\textwidth]{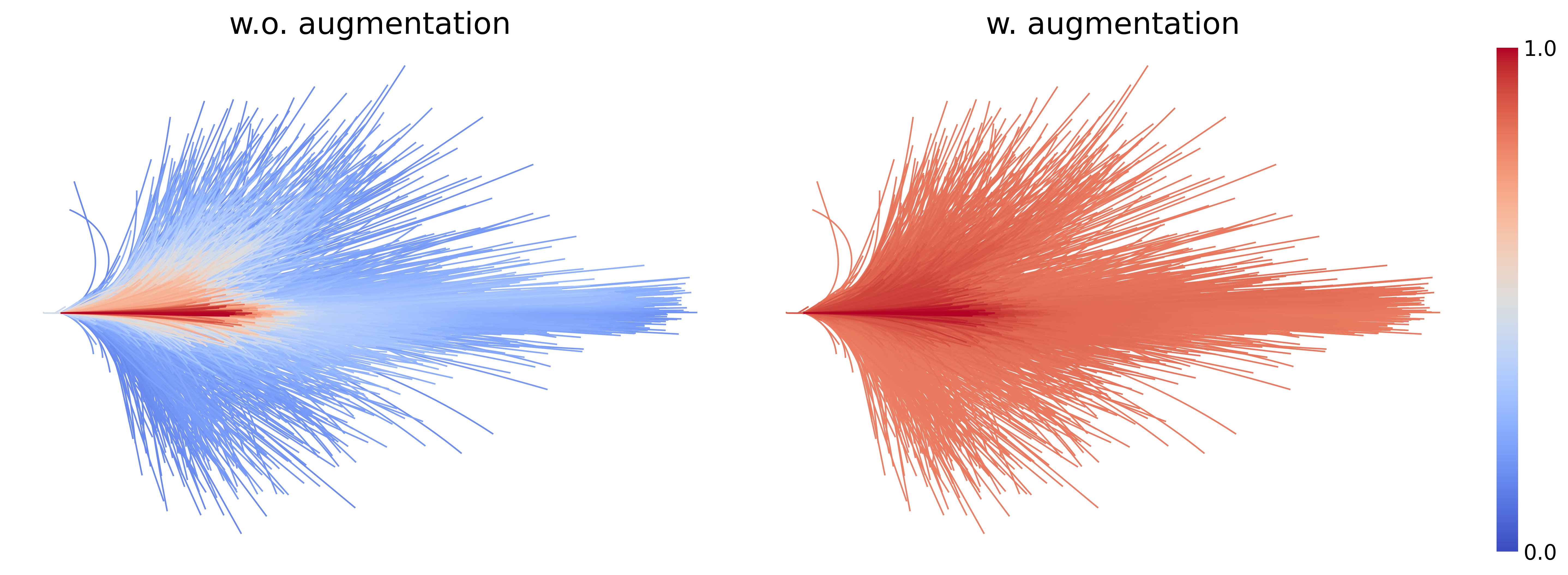}
    \caption{Comparison of dataset high-score trajectory distribution. Our augmentation mitigates directional bias, significantly increasing the frequency of previously underrepresented turning trajectories. The color bar represents the normalized frequency of different trajectories in the dataset. } \label{fig:trajectory_cmp}
\end{figure}

\paragraph{Trajectory distribution comparison}
Fig.~\ref{fig:trajectory_cmp} illustrates the trajectory frequency distribution in both the original dataset and the dataset enhanced with our rotation-based augmentation method. Using the ground-truth PDM score, we count the trajectory that either scores above 0.99 or ranks among the top-three-highest scores in each scenario. We then normalize the frequency by setting the most frequent trajectory to 1. Results show that in the original dataset, the high-score trajectories are predominantly concentrated in the forward or near-forward direction, whereas in the augmented dataset, trajectories across all directions appear with similar frequency.


\subsection{E. Limitations and Future Work}
\label{sec:supp_limitations}
In this paper, we evaluate the performance of \modelName on the NAVSIM and Bench2Drive benchmark. Our two-stage coarse-to-fine trajectory filtering approach proves effective in enhancing model performance. However, extending this filtering strategy to a multi-stage setting does not yield additional improvements, and the model still has potential to improve performance under challenging or corner cases.
In the future, we will further investigate multi-stage refinement techniques and explore reinforcement learning for a more robust planning system.

\section{Acknowledgments}
This work was supported by National Natural Science Foundation of China (No. 62427819) and the Science and Technology Commission of Shanghai Municipality (No. 24511103100).

\newpage

\end{document}